\documentclass[runningheads,11pt]{llncs}
\usepackage{float}
\usepackage{booktabs}
\usepackage{graphicx}
\usepackage{amsmath,amssymb}
\usepackage{hyperref}
\usepackage{multirow}
\usepackage{xcolor}
\usepackage[margin=2.5cm]{geometry}
\usepackage{setspace}
\title{Through the Gaps: Uncovering Tactical Line-Breaking Passes with Clustering}

\author{
Oktay Karakuş\inst{1} \and
Hasan Arkadaş\inst{2}
}

\institute{
Cardiff University, School of Computer Science and Informatics, UK\\
\email{karakuso@cardiff.ac.uk}
\and
Independent Researcher \\
\email{hasan@deadball-analytics.com}
}
\onehalfspace
\begin{document}
\maketitle

\begin{abstract}
Line-breaking passes (LBPs) are crucial tactical actions in football, allowing teams to penetrate defensive lines and access high-value spaces. In this study, we present an unsupervised, clustering-based framework for detecting and analysing LBPs using synchronised event and tracking data from elite matches. Our approach models opponent team shape through vertical spatial segmentation and identifies passes that disrupt defensive lines within open play. Beyond detection, we introduce several tactical metrics, including the space build-up ratio (SBR) and two chain-based variants, LBPCh$^1$ and LBPCh$^2$, which quantify the effectiveness of LBPs in generating immediate or sustained attacking threats. We evaluate these metrics across teams and players in the 2022 FIFA World Cup, revealing stylistic differences in vertical progression and structural disruption. The proposed methodology is explainable, scalable, and directly applicable to modern performance analysis and scouting workflows.

\end{abstract}

\section{Introduction}
Over the past decade, football analytics has been transformed by the rise of data science and the availability of high-resolution spatiotemporal data. This evolution has enabled analysts and researchers to move beyond basic statistics toward deeper tactical insights. Seminal contributions like Expected Goals (xG) \cite{lucey2015quality,eggels2016explaining,hewitt2023machine,scholtes2024bayes} and pass valuation frameworks \cite{bransen2019measuring,gyarmati2014qpass,decroos2019actions} have highlighted the importance of space, movement, and context in shaping match outcomes.

Within this landscape, line-breaking passes (LBPs) have emerged as a key tactical mechanism, enabling teams to penetrate opponent defensive or midfield lines and access advanced attacking zones. These passes often lead to goal-scoring opportunities by disrupting spatial structure and eliminating multiple defenders through mostly vertical play \cite{gyarmati2014qpass,tayyab2024space}. Despite their importance, identifying LBPs from raw data remains a nontrivial task. Definitions vary across commercial platforms and academic literature, often relying on proprietary labels or manual annotations \cite{andrienko2017visual,decroos2019actions,optavision2023}. Moreover, many existing methods lack interpretability and do not explicitly account for the opponent’s team structure.

In this paper, we introduce an unsupervised, data-driven framework to detect and quantify LBPs using a clustering-based model of defensive structure. Leveraging synchronised event and tracking data from the 2022 FIFA World Cup, our method segments opponent team shape into vertical defensive bands and identifies passes that disrupt these formations. Unlike approaches reliant on predefined formations or proprietary labels, our model infers spatial disruption directly from positional configurations at the time of the pass.

Beyond detection, we propose novel tactical metrics to evaluate the quality and intent of LBPs, including the Space Build-up Ratio (SBR), LBPCh$^1$ (a single LBP leading to a shot), and LBPCh$^2$ (linked LBPs culminating in a goal attempt). These measures enable rich team and player profiling of verticality and offensive structure. Our framework is interpretable, reproducible, and scalable, making it suitable for both tactical analysis and scouting applications, and laying the groundwork for future learning-based extensions.

\section{Related Work}
In football analytics, a key distinction exists between \textit{progressive passes}, which advance the ball significantly toward the opponent’s goal (typically by 10–30 meters~\cite{wyscout}), and \textit{line-breaking passes} (LBPs), which explicitly penetrate structured lines of opposing players~\cite{yorke2022}. While progressive passes capture directional intent, they do not account for defensive positioning. LBPs, by contrast, reflect a team’s ability to disrupt shape and bypass opponents vertically. Their tactical value has been widely acknowledged across both academic and applied domains. Foundational work such as \cite{gyarmati2014qpass} introduced pass value frameworks based on ball movement and team contribution, while follow-up studies incorporated spatial awareness and defensive structure~\cite{szczepanski2016beyond,bransen2017valuing}. Michalczyk~\cite{michalczyk2020} further showed that LBPs double the likelihood of leading to a goal compared to other passes.

Identifying LBPs from data is challenging due to the need to model the opponent’s shape. Some rule-based methods rely on tracking data to identify defensive lines and geometrically evaluate whether a pass intersects them~\cite{michalczyk2020}. StatsBomb’s 360 approach~\cite{yorke2022} uses a combination of positional data and heuristics such as forward movement and relative position of defenders to classify LBPs. Recent efforts have focused specifically on identifying and characterising LBPs via learning-based methods, for example, Michalczyk~\cite{michalczyk2020} proposed geometric heuristics based on line intersection to classify passes, while \cite{el_kadi2024ml} experimented with supervised learning models trained on annotated examples. Industry systems such as StatsBomb 360° \cite{yorke2022} and Driblab’s Arrigo \cite{driblab2025arrigo} offer proprietary methods for detecting line-breaking actions using freeze-frame contextual data. However, these systems often rely on predefined formations or third-party annotations, limiting interpretability and reproducibility.

The accurate representation of a defensive organisation is central to identifying LBPs. Early approaches used spatiotemporal clustering to infer team formations~\cite{bialkowski2014}. More recent work captures dynamic shape: Narizuka and Yamazaki~\cite{narizuka2019} applied hierarchical clustering to connectivity graphs built from Delaunay triangulations of player positions. Others model team compactness using geometric descriptors like convex hulls~\cite{memmert2017networks}, or spatial dispersion measures such as inter-line distances. Zardiny and Bahramian~\cite{zardiny2025} proposed a clustering framework that extracts possession-specific shape patterns based on spatial and geometric features.

Several metrics aim to quantify the impact of actions on team shape. The ``packing'' metric~\cite{packing} remains a popular benchmark for assessing how many defenders are bypassed by a pass. Others have explored changes in convex hull size or shape symmetry to capture disruption. Possession-value models such as Expected Threat (xT)~\cite{singh2018}, Valuing Actions by Estimating Probabilities (VAEP)~\cite{van2020valuing,decroos2021vaep} or On-Ball Value (OBV)~\cite{statsbomb_obv} implicitly reward LBPs by assigning higher value to passes entering dangerous zones.

\vspace{0.2cm}\noindent\textbf{Relation to our approach.} While prior work has addressed LBP detection, team shape modelling, and structural metrics independently, our approach combines these strands in a unified, unsupervised framework by clustering opponent positions and detecting penetrations. We quantify vertical disruption directly from spatiotemporal data without supervision, and this interpretability allows for deeper tactical insights into player-team level vertical progression.

\section{Datasets}

We use the publicly available 2022 FIFA World Cup dataset released by PFF FC\footnote{\url{https://www.blog.fc.pff.com/blog/pff-fc-release-2022-world-cup-data}}, which provides synchronised event and tracking data for all 64 matches. This enables detailed spatiotemporal analysis of player behaviour, team structure, and ball progression. Each match includes structured JSON files containing: \textit{event data} (timestamped on-ball actions with spatial and contextual tags), \textit{tracking data} (29.97Hz player and ball positions), \textit{metadata} (pitch dimensions, orientation, frame rate), and \textit{roster data} (linking jersey numbers to player IDs and roles). We use smoothed tracking positions to reduce jitter. For example, the Senegal vs Netherlands match contains 1,800+ events and 160,000 tracking frames. This rich context forms the foundation for our unsupervised line-breaking pass detection framework.

\section{Methodology}
Our methodology aims to detect and quantify LBPs using a clustering-based model of opponent team structure. We also introduce tactical metrics such as SBR and LBP chains to evaluate their spatial and strategic impact.

\subsection{Line-Breaking Pass Detection via Clustering}
We model the opponent team shape using vertical segmentation. At the time of each pass, opponent player positions are grouped into $k$ clusters based on their lateral ($x$-axis) coordinates using agglomerative clustering \cite{mullner2011modern}. This approach groups players into vertical bands approximating tactical lines (e.g., a back four or midfield block).

Let a pass $p_i$ occur at time $t_i$, with passer position $s_i = (x_s, y_s)$ and receiver position $r_i = (x_r, y_r)$. Let $\{C_j\}_{j=1}^k$ denote the set of vertical opponent clusters at $t_i$, where each cluster $C_j$ is represented by a vertical segment defined by a horizontal centroid $x_j$, and a vertical span $[y_{\min}^{(j)}, y_{\max}^{(j)}]$, based on the minimum and maximum $y$-coordinates of players in $C_j$. We define a pass as line-breaking under the clustering model if it crosses the $x$-centroid of any cluster and intersects the vertical segment defined by that cluster's lateral span. Formally:
\begin{equation}
\text{LBP}_{\text{cluster}}(p_i) = \mathbb{I} \left( \exists C_j : x_j \in (x_s, x_r) \land \text{SegmentIntersects}(p_i, C_j) \right)
\end{equation}

\noindent where $\text{SegmentIntersects}(p_i, C_j)$ is true if the path from $s_i$ to $r_i$ intersects the vertical segment at $x = x_j$, bounded by $[y_{\min}^{(j)}, y_{\max}^{(j)}]$. 
In addition to structural intersection, we apply two supporting filters: (1) the pass must bypass at least two opponents, based on their lateral proximity to the pass vector, and (2) the pass must be forward, relative to team orientation.

\vspace{0.1cm}\noindent\textbf{(LBP Volume)} The total number of LBPs per team or player serves as a baseline measure of verticality and structural aggression. Frequent LBPs suggest proactive build-up play, tactical sharpness, or positional superiority. At the player level, they highlight individuals who consistently attempt to break lines, often midfielders or wide players inverting centrally. 

\vspace{0.1cm}\noindent\textbf{(Direct Vertical Threat, LBPCh$^1$)} To measure the immediate impact of LBPs, we define \textit{LBPCh$^1$} as the subset of LBPs that directly lead to a shot or an assist within the same possession. These actions indicate a direct vertical threat, with little delay or buildup between structural disruption and goal creating chances. High LBPCh$^1$ values reflect players/teams capable of fast, incisive play after breaking lines.

\vspace{0.1cm}\noindent\textbf{(Sustained Vertical Progression, LBPCh$^2$)} In contrast to immediate threat, \textit{LBPCh$^2$} measures sustained tactical progression, and identifies cases where one LBP is immediately followed by another within the same team possession. This sequence then concludes with a shot/assist. It captures structurally coherent attacks that maintain vertical momentum across multiple passes, often revealing more collective and layered approaches to play.

\subsection{Space Build-Up Ratio (SBR)}

\begin{figure}[ht]
    \centering
    \includegraphics[width=\textwidth]{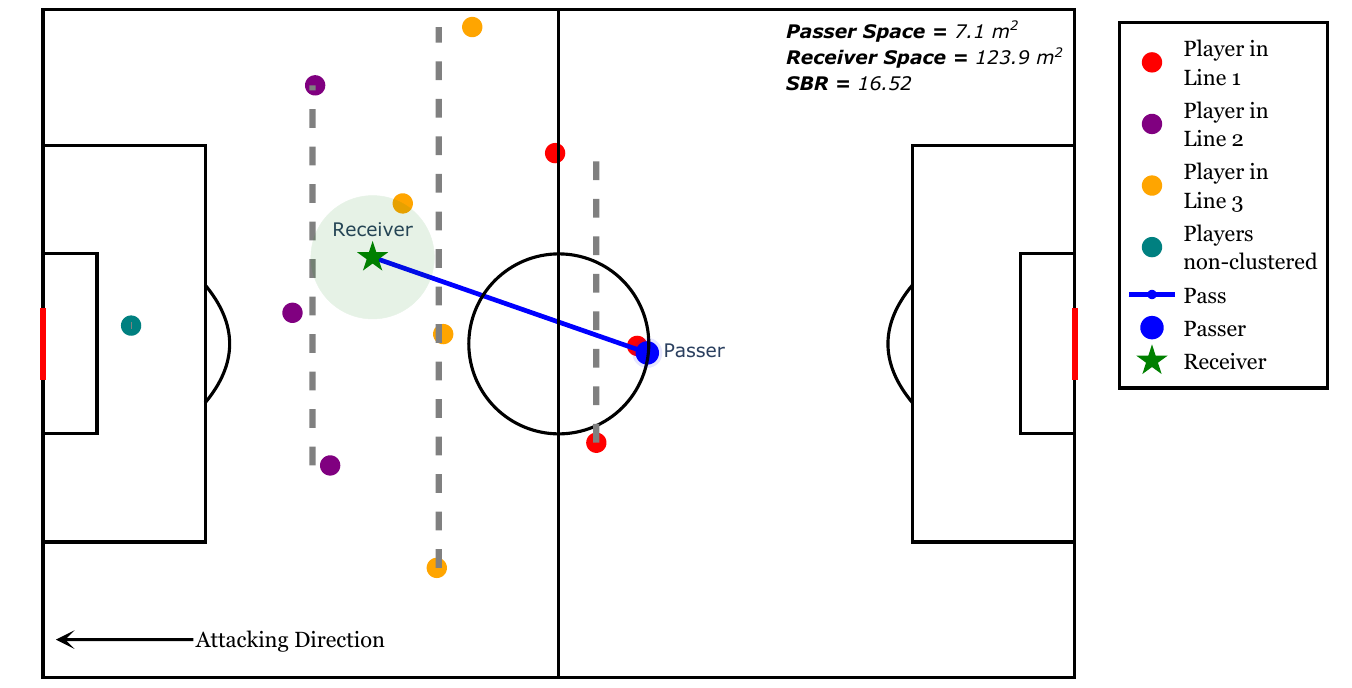}
    \caption{Illustration of an LBP and an SBR.}
    \label{fig:example_lbp}
\end{figure}

To complement structural disruption, we define the \textit{SBR} as a spatial metric that quantifies whether a pass moves the ball into a less congested area. Let $A_p = \pi d_p^2$ and $A_r = \pi d_r^2$ be circular estimates of space around the passer and receiver based on the distance to their nearest opponent. Hence, SBR is:
\begin{equation}
\text{SBR}(p_i) = \frac{A_r - A_p}{A_p} = \frac{d_r^2}{d_p^2} - 1.
\end{equation}
A positive SBR indicates that the receiver is under less pressure than the passer, interpreted as \textit{space opened up}. A negative value reflects an increase in defensive pressure. This metric is used to compare the spatial gain of different LBPs and can be related to downstream outcomes such as assists or shots. Figure~\ref{fig:example_lbp} illustrates a detected LBP with an annotated SBR. Opponent players form three visually structured defensive layers, suggesting a 3-4-3 formation. The pass breaks both the forward and midfield vertical lines, reaching a receiver (green star) positioned in a higher-space region, with an SBR of 16.52. Semi-transparent circles represent the local area around each player.

\section{Experimental Evaluation}
\begin{figure}[ht!]
    \centering
    \includegraphics[width=\textwidth]{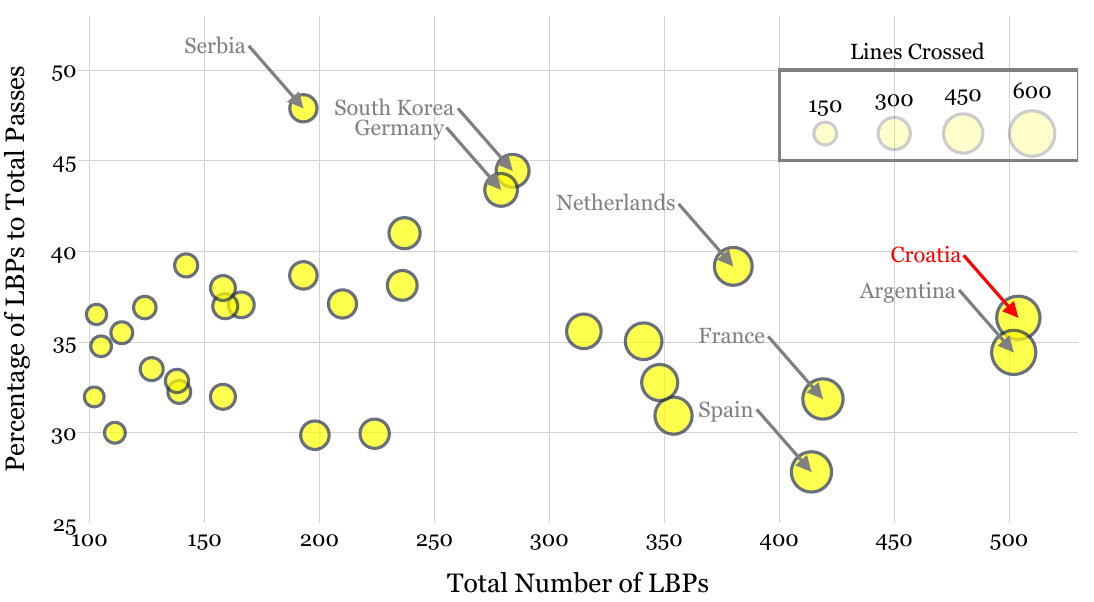}\\\vspace{1.5cm}\includegraphics[width=\textwidth]{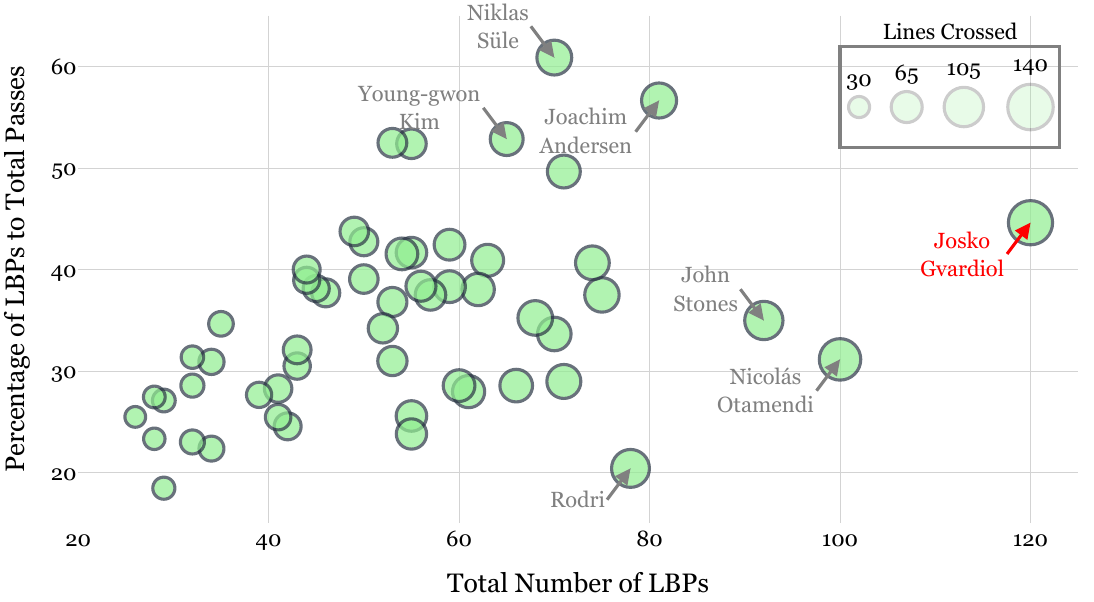}
    \caption{Most frequent line-breakers as detected by our clustering-based model. \textbf{Top}: Team-level; \textbf{Bottom}: Player-level. Each bubble represents an entity about LBPs where the \textit{bubble size} reflects the total number of defensive lines broken. 
    }
    \label{fig:LBP_res}
\end{figure}

We begin our analysis by examining which teams and players most frequently attempt to break opponent structures through LBPs. Unlike simple pass counts, LBP volume highlights how often a team or player chooses to disrupt vertical opponent lines, a critical marker of tactical intent and progressive play. Across all 64 matches and 21,349 passes in the dataset, the proposed model identified 7,477 LBPs, averaging around 117 per match. Figure~\ref{fig:LBP_res} presents teams and players ranked by total number of LBPs over the tournament.

From a team perspective, a high volume of LBPs reflects a deliberate tactical commitment to progressing through structured defensive lines rather than bypassing them via wide areas. Possession-oriented sides such as Croatia, Argentina, France, and Spain dominate the rankings, with Serbia also notable for frequent line-breaking despite a lower overall volume. These teams consistently seek central occupation and third-man combinations, principles aligned with modern positional play. Notably, three of the four semi-finalists rank among the top teams in total LBP count, underscoring the strategic value of vertical penetration in elite tournament success.

At the player level, midfielders and fullbacks emerge as primary line-breakers often responsible for initiating build-up, operating between lines, and delivering progression into advanced zones. Standout contributors such as Gvardiol, Rodri, and Otamendi exemplify these roles, functioning as deep facilitators who not only disrupt opposition shape but also help maintain compact defensive structure during transitions. Their involvement in LBPs highlights the blend of technical precision and tactical intelligence required to consistently play through pressure.

\begin{figure}[ht!]
    \centering
    \includegraphics[width=\textwidth]{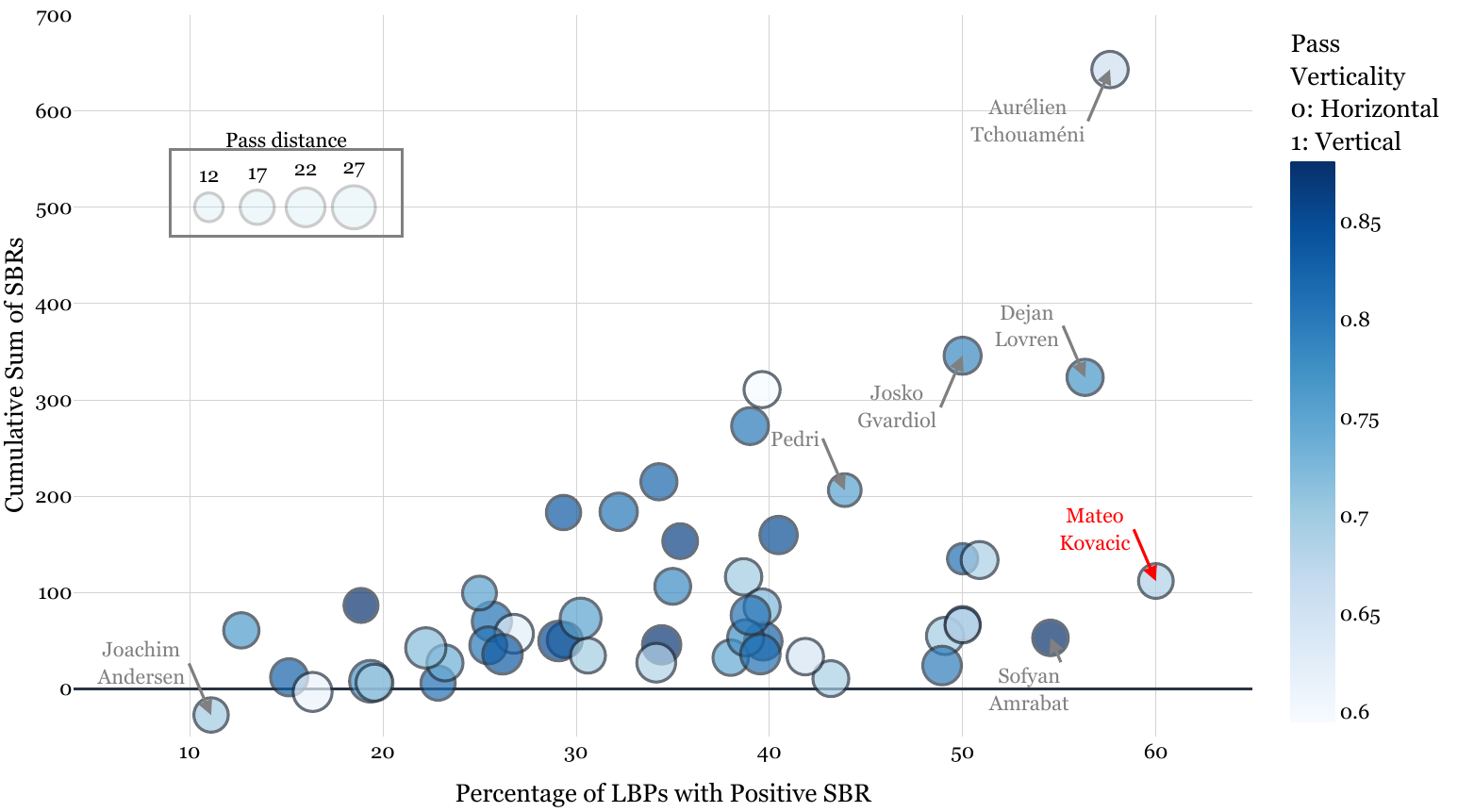}
    \caption{Space progression effectiveness of the top 50 players. The x-axis shows the percentage of LBPs with positive SBR, whilst the y-axis represents the cumulative SBR. Bubble size indicates average pass distance, and colour encodes pass verticality.}
    \label{fig:sbr}
\end{figure}

\subsection{Spatial Impact: The Role of SBR}

To assess the spatial effectiveness of LBPs, we plot cumulative SBR against the percentage of LBPs yielding positive SBR values, focusing on the top 50 players by LBP count. In Figure~\ref{fig:sbr}, bubble size indicates average pass distance, while colour denotes pass verticality. This visualisation reveals distinct tactical profiles: players like Tchouaméni and Gvardiol, in the upper/mid right region, consistently generate space and accumulate high total gains. In contrast, Amrabat shows a high success rate but modest cumulative values despite topping the verticality scale, reflecting shorter controlled disruptions.

The framework also exposes cases where volume alone is misleading. Despite his high LBP count shown in Figure \ref{fig:LBP_res}, Andersen registers low cumulative SBR and few space-generating passes, showing that frequent LBPs does not guarantee tactical value. In general, SBR offers a more nuanced and unsupervised measure of progression. Unlike pass distance, which can misrepresent impact, SBR captures the receiver’s spatial advantage, helping distinguish between merely long passes and truly effective ones. It is important to note that all top 50 players in this analysis exhibit verticality values above 0.6, underscoring the strong association between LBPs and vertical tactical build-up.

\subsection{From Break to Threat: Direct Line-Break to Chance (LBPCh$^1$)}
While LBPs inherently reflect verticality and structural disruption, not all such passes culminate in tangible offensive rewards. To assess the immediate impact of LBP, we define LBPCh$^1$ as the subset of LBP that directly precedes a shot/assist within the same possession phase. 

\begin{figure}[ht!]
    \centering
    \includegraphics[width=0.9\textwidth]{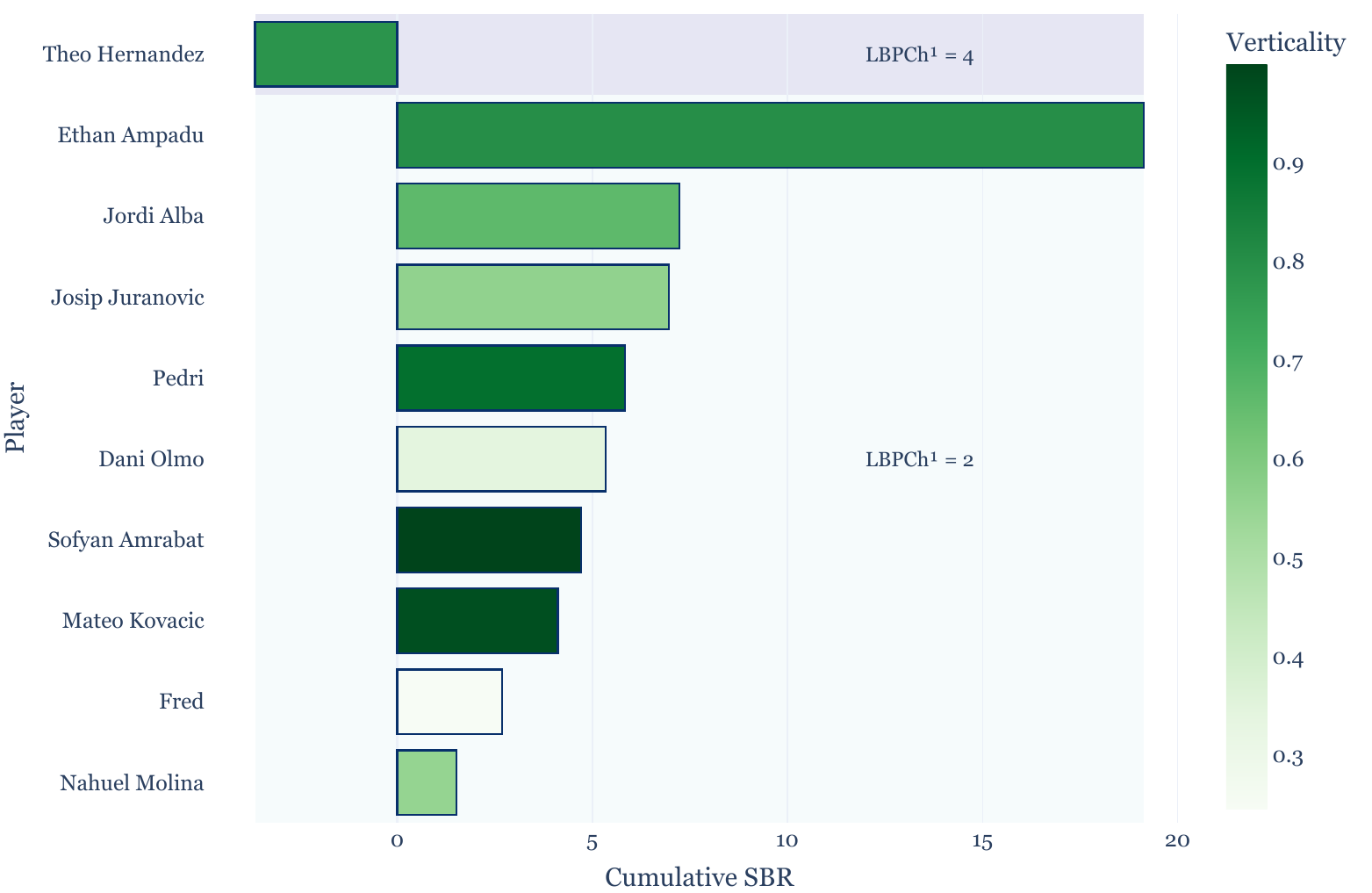}\\\vspace{0.5cm}\includegraphics[width=0.9\textwidth]{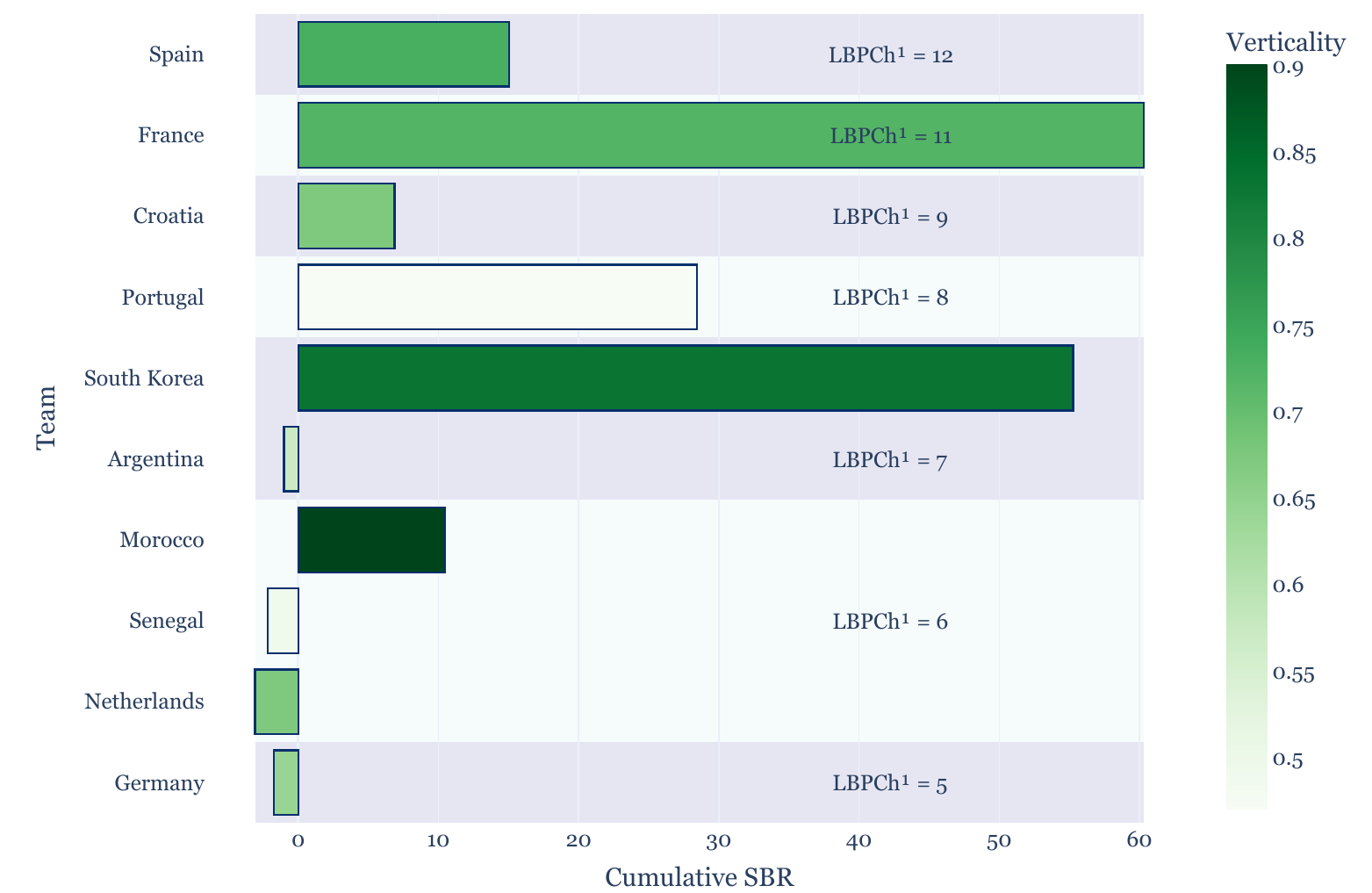}
    \caption{Direct Line-Breaks to Chances (LBPCh$^1$). Team (Top) and Player (Bottom) Level. Bars (i) represent the cumulative SBR for LBPs that directly led to a shot, (ii) are coloured by average pass verticality, and (iii) are ordered by total LBPCh$^1$. 
    }
    \label{fig:lbpch1}
\end{figure}

Figure \ref{fig:lbpch1} presents team/player-level summaries of LBPCh$^1$ instances. At the team level (Fig. \ref{fig:lbpch1}-Bottom), traditional possession-heavy sides such as Spain, France, Croatia and Portugal lead the total LBPCh$^1$ events, combining high structural disruption with immediate offensive conversion. Interestingly, Morocco and South Korea also appear in the top ranks, highlighting their efficient use of direct vertical sequences, often built through transitional attacks rather than prolonged build-up. From a player perspective (Fig. \ref{fig:lbpch1}-Top), Theo Hernández stands out as the only player in the dataset with four LBPCh$^1$ events. His performance underlines the value of progressive fullbacks in modern systems, able to carry or pass through pressure and deliver directly threatening balls into the final third. A cluster of players, including Pedri, Amrabat, and Fred, follow closely with two instances each. These midfielders are known for their positional awareness and capacity to operate between lines, often acting as the link between phases of possession and final-third penetrations.

The plots incorporate both cumulative SBR and pass verticality to provide deeper tactical context. Higher cumulative SBR values suggest that not only were these passes progressive, but they also opened up significant space for receivers. For example, Jordi Alba and Juranovic deliver LBPCh$^1$ passes with relatively lower verticality but high SBR, indicating subtle through balls into expanding pockets rather than long direct deliveries. In contrast, Amrabat, Pedri and Kovacic exhibit high verticality with shorter build-ups, likely reflecting more abrupt, transitional progressions from deeper areas.


\subsection{Sustained Threat: LBP Chains Leading to Chances (LBPCh$^2$)}
While LBPCh$^1$ reflect individual tactical sharpness or isolated exploitation of space, longer chains of connected LBPs culminating in shots offer deeper insight into coordinated vertical build-up. To evaluate this sustained threat, we introduce LBPCh$^2$, defined as two consecutive LBPs within the same possession phase that result in a shot, goal, or assist.

\begin{figure}[ht]
    \centering
    \includegraphics[width=\textwidth]{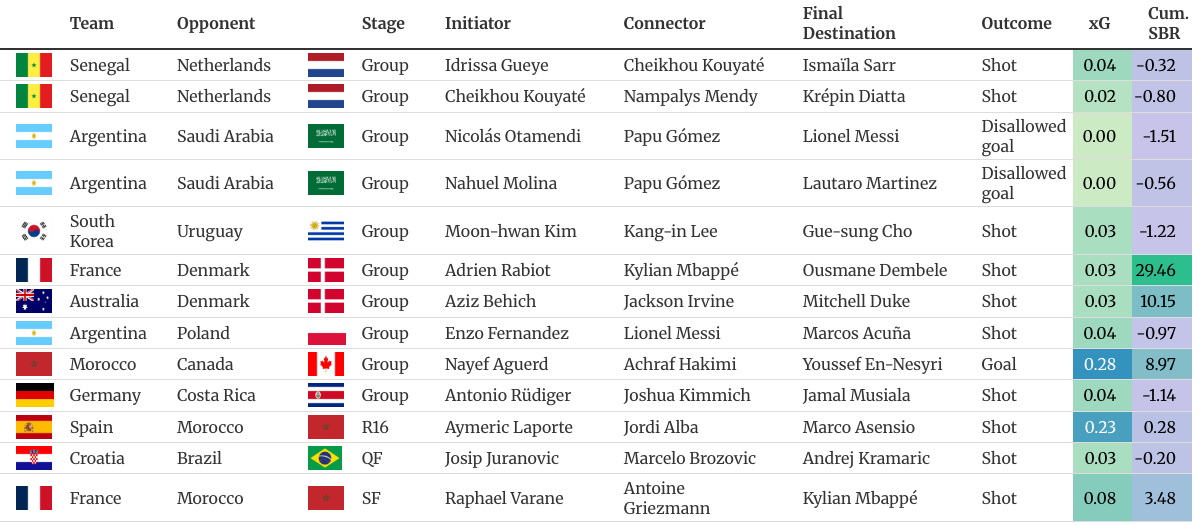}
    \caption{Summary of all detected LBPCh$^2$ sequences. Each row represents an LBPCh$^2$. The table includes the initiating, connecting, and finishing players, match details, final outcomes (with xG) and chain cumulative SBR values.}

    \label{fig:lbpch2}
\end{figure}

Figure~\ref{fig:lbpch2} summarises the 13 such sequences detected in the entire tournament. These multi-pass sequences, though rare, signal the capacity to construct coherent progression through compact opponent formations. From a team-level perspective, Argentina emerges as a notable case. Despite two of their three LBPCh$^2$ sequences being ruled offside by VAR, they reveal clear tactical coordination. In both cases, Papu Gómez acted as the connector, receiving the first LBP and immediately delivering a second. This repetition not only indicates structured build-up but also points to Argentina’s midfield balance, where wide creators like Gómez operate between lines. 
France, Germany, and Spain also feature, demonstrating that top sides are capable of stringing vertical actions in phases.
Only one chain in the data set resulted in a goal: the Morocco sequence involving Aguerd-Hakimi-En-Nesyri, which illustrates how LBPCh$^2$ can serve as a direct scoring route even for defensively focused teams.


On the player level, the figure highlights key roles within the chain: \textit{Initiators} like Enzo Fernández, Josip Juranović, and Adrien Rabiot consistently make the first vertical incision. \textit{Connectors} such as Cheikhou Kouyaté and Achraf Hakimi provide continuity, either through turns or quick layoffs. \textit{Finishers} include elite attackers like Ousmane Dembele, Lionel Messi and Kylian Mbappé.
Interestingly, xG values across these sequences remain mostly below 0.10. This supports the notion that LBPCh$^2$ sequences often occur under spatial pressure, requiring multiple actions to unlock central zones. High cumulative SBRs and verticality are common, reflecting the nature of these chain-progressions as strategic rather than opportunistic.


\section{Conclusion}
In this study, we proposed an unsupervised, clustering-based framework to detect and analyse LBPs using synchronised event and tracking data. By modelling defensive structure through dynamically segmented vertical clusters, we evaluated each pass in terms of structural disruption and spatial progression. Our formulation introduced the SBR as a novel spatial metric, enabling a data-driven distinction between merely vertical and truly effective passes.

The evaluation across all 64 matches of the 2022 FIFA World Cup uncovered meaningful tactical signals. LBPs were strongly associated with teams reaching the later stages of the tournament, with fullbacks and deep-lying midfielders emerging as key contributors. Through SBR analysis, we highlighted player-level variability in generating spatial advantage, revealing differences not captured by volume alone.

We further introduced LBPCh$^1$ and LBPCh$^2$ to measure how structural disruption translates into attacking output. While LBPCh$^1$ captured direct vertical threat, LBPCh$^2$ revealed sustained progression patterns involving coordinated build-up. These metrics collectively offer a new lens for scouting, tactical analysis, and performance diagnostics.

Future work includes learning-based models that could extend the proposed framework, using the clustering approach as a pretext task or interpretable benchmark. By releasing the code and curated metrics, we aim to contribute toward transparent, scalable tools for spatiotemporal football analytics.

\bibliographystyle{splncs04}
\bibliography{refs}

\end{document}